\documentclass[conference]{IEEEtran}
\IEEEoverridecommandlockouts
\usepackage{cite}
\usepackage{amsmath,amssymb,amsfonts}
\usepackage{algorithm}
\usepackage{algorithmic}
\usepackage{graphicx}
\usepackage{textcomp}

\usepackage{booktabs} % Allows the use of \toprule, \midrule and \bottomrule in tables
\usepackage{multirow}
\usepackage{float}
\usepackage{graphicx}
\usepackage{subfigure}
\usepackage{enumerate}
\usepackage{color}

\def\BibTeX{{\rm B\kern-.05em{\sc i\kern-.025em b}\kern-.08em
    T\kern-.1667em\lower.7ex\hbox{E}\kern-.125emX}}
\begin{document}

\title{Performance Optimization of a Fuzzy Entropy based Feature Selection and Classification Framework\\
	%An Optimized Framework for Fuzzy Entropy based Feature Selection and Classification\\
% {\footnotesize \textsuperscript{}}
}

\author{\IEEEauthorblockN{Zixiao Shen, Xin Chen, Jonathan M. Garibaldi}
\IEEEauthorblockA{\textit{Intelligent Modelling and Analysis Group, School of Computer Science} \\
\textit{University of Nottingham}, Nottingham, NG8 1BB, UK \\
\{Zixiao.Shen, Xin.Chen, Jon.Garibaldi\}@nottingham.ac.uk}
}

\IEEEspecialpapernotice{IEEE International Conference on Systems, Man, and Cybernetics (SMC), 2018, Miyazaki, Japan}

\maketitle

\begin{abstract}
In this paper, based on a fuzzy entropy feature selection framework, different methods have been implemented and compared to improve the key components of the framework. Those methods include the combinations of three ideal vector calculations, three maximal similarity classifiers and three fuzzy entropy functions. Different feature removal orders based on the fuzzy entropy values were also compared. The proposed method was evaluated on three publicly available biomedical datasets. From the experiments, we concluded the optimized combination of the ideal vector, similarity classifier and fuzzy entropy function for feature selection. The optimized framework was also compared with other six classical filter-based feature selection methods. The proposed method was ranked as one of the top performers together with the Correlation and ReliefF methods. More importantly, the proposed method achieved the most stable performance for all three datasets when the features being gradually removed. This indicates a better feature ranking performance than the other compared methods.
\end{abstract}

%\begin{IEEEkeywords}
%component, formatting, style, styling, insert
%\end{IEEEkeywords}

\section{Introduction}
Due to the rapid development and wide application of information technology, increasing amount of data with rich information are generated. Discovering the information that concealed in these datasets becomes essential and challenging. In real world applications, datasets often contain irrelevant and redundant features that do not provide useful or additional information for subsequent decision makings \cite{kumar2014feature}. According to Occam's razor, it is important and necessary to eliminate those irrelevant and redundant features \cite{domingos1999role}. Therefore, feature selection has always been an active research area, particularly when more and more 'big data' become available in many application areas.

Recently, machine learning methods have achieved superior performance in many application areas such as diagnostic decision making and disease classification \cite{kononenko2001machine}. However, the high dimensionality and complexity of the data often make the methods suffering from the problem of curse of dimensionality \cite{bellman2013dynamic}. With the high dimensionality, the limited number of training samples are sparsely distributed in the feature space. This makes the machine learning methods difficult to learn the underlying relationship accurately. This also leads to an over-fitting problem that the learned model is not generic enough for unseen data samples. In addition, high dimensional datasets significantly increase the memory usage and computational cost for data analysis, which result in low efficiency of algorithms \cite{li2016feature}.

Dimensionality reduction methods aim to deal with the aforementioned issues, which are mainly classified into feature extraction and feature selection methods. The feature extraction methods tend to project the original high dimensional feature into a new low dimensional feature space, where the contributions of features are combined. In contrast, feature selection methods maintain a subset of the original features that are highly relevant to the subsequent decision makings \cite{li2016feature}. This provides the model with better readability and interpretability, which are essential for our main application area (biomedical dataset). Our proposed method belongs to the feature selection category.

Practical biomedical datasets are usually imperfect which contain uncertain texts and incomplete features. The uncertainty will increase after applying some data analysis processes \cite{bandemer2012fuzzy}. Fuzzy logic algorithms are designed to model the vagueness, imprecision and uncertainty. In order to overcome the practical problems embedded in the biomedical datasets, it becomes a natural choice to integrate fuzzy logic methods into the feature selection process.

Various fuzzy methods have been proposed for feature selection. In 1999, Rezaee \cite{rezaee1999fuzzy} presented a method to automatically identify the reduced linguistic fuzzy set of a labeled multi-dimensional dataset. The optimal subset of fuzzy features is determined by projecting the original data set onto a fuzzy space. In 2002, Rui-Ping Li \cite{li2002fuzzy} proposed a fuzzy neural network method for pattern classification and feature selection. The proposed neural network attempts to select the important features from the original features while maintaining the maximum recognition rate. In 2008, Tsang \cite{tsang2008attributes} introduced a concept of attributes reduction with fuzzy rough sets and developed an algorithm using discernibility matrix to compute all the attributes reductions. More recently, Luukka \cite{luukka2011feature} introduced a fuzzy entropy feature selection framework based on a maximal similarity classifier. The framework is computationally efficient, readily comprehensible and easily adapted to different applications compared with other fuzzy feature selection methods. The original framework consists of three fundamental components, namely ideal vector calculation, similarity measurement and fuzzy entropy calculation. Different measurements can be used in these components, which affect the feature selection performance. To the authors' best knowledge, a comprehensive comparison using different measurements within these components has not been reported, so as the comparison to other feature selection methods.

In this paper, we implemented three different measurements for each of the key components in the framework and comprehensively compared the performances based on different combinations of the measures. Additionally, it is also compared with other six feature selection methods in the literature. All the evaluations were tested based on three widely used publicly available biomedical datasets \cite{Lichman2013}.

%%-------------------------------------------------------------------------------------------------------------------------------------------------------------------------------------------------------------------
\section{Methodology}
\begin{figure}[h]
	\centering
	\includegraphics[width=0.45\textwidth]{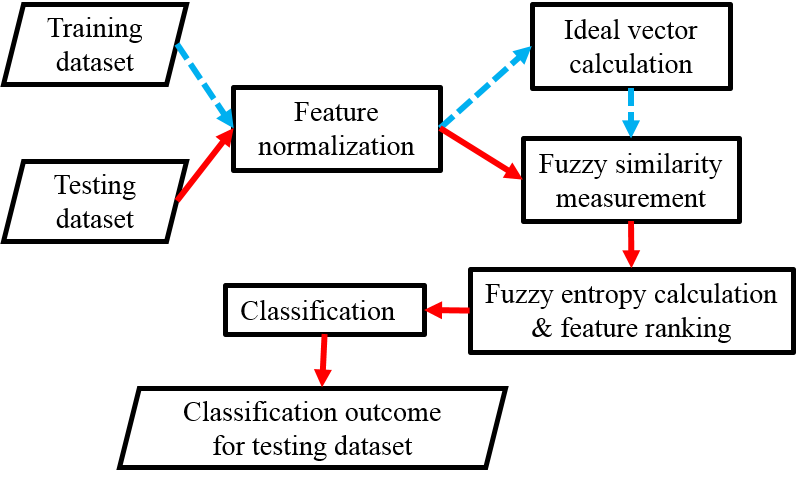}
	\caption[Fuzzy entropy based feature selection flowchart]
	{Overview of the proposed framework. Blue dotted line and red solid line are the data flows for training and testing processes respectively.}
	\label{flowchartFS}
\end{figure}

Based on the method in \cite{luukka2011feature}, we proposed a data driven framework to deal with feature selection and classification. The overall structure is illustrated in Fig. \ref{flowchartFS}.

The proposed method aims to classify a total number of $M$ subjects into $N$ different classes $C_k,  k\in[1, N]$ by their feature vector $\vec{x_i}$. $i$ is the index of the subjects, and the number of features in $\vec{x_i}$ is denoted by D. The procedure of the proposed method is described as follows.

\begin{enumerate}[Step 1:]
	\item For the training set, normalize each feature value to the range of [0 1] using min-max normalization method \cite{jain2011min}. The maximum value of each feature needs to be carefully determined to avoid using outliers and consistently applied to both training and testing datasets.
	\item Based on the normalized values from Step 1, calculate the ideal vector $\vec{v}_k$ for the $k^{th}$ class.
	\item Apply the same normalization process in Step 1 to the testing set.
	\item Calculate the fuzzy similarity between the feature vector $\vec{x_i}$ of the testing subjects and the ideal vector $\vec{v}_k$ obtained in Step 2.
	\item Based on the similarity values in Step 4, construct a $MN \times D$ similarity matrix. Calculate the fuzzy entropy value of each feature (column of the matrix) and subsequently rank the features according to the values.
	\item Select the features and classify the testing set based on the ranked feature sequence from Step 5.
\end{enumerate}

The detailed descriptions of ideal vector calculation, similarity measurement, fuzzy entropy calculation and classification are given in the following subsections.

\subsection{\textbf{Ideal vector calculation based on training dataset}}\label{Idealvector}
Ideal vector is used to represent the "mean" property of the subjects in each class. Different methods of calculating the ideal vector, namely arithmetic mean, geometric mean and harmonic mean have been implemented and compared as follows. In equations (1-3), $N_k$ is the number of subjects for the $k^{th}$ class. $j$ is the index of features. The remaining notations are the same as previously introduced. The performance comparing different ideal vector calculations is reported in section \ref{expCombinations}.

\paragraph{Arithmetic mean}\label{eqAm}
\begin{equation}
\vec{v}^A_k(j) = \frac{\sum_{i=1}^{N_k}\vec{x_i}(j)}{N_k}, \  j \in [1,D]
\end{equation}

\paragraph{Geometric mean}\label{eqGm}
\begin{equation}
\vec{v}^G_k(j) = \sqrt[N_k]{\prod_{i=1}^{N_k}\vec{x_i}(j)}, \  j \in [1,D]
\end{equation}

\paragraph{Harmonic mean}\label{eqHm} 
\begin{equation}
\vec{v}^H_k(j) = \frac{N_k}{\sum_{i=1}^{N_k} \frac{1}{\vec{x_i}(j)}}, \  j \in [1,D]
\end{equation}

\subsection{\textbf{Fuzzy similarity measurement}}\label{SimilarityMeasures}

In this section, the similarity measurement is presented in the form of generalized $L$ukasiewicz algebra \cite{saastamoinen2003testing}. It measures the similarity between the $j^{th}$ element of feature vector $\vec{x_i}$ and the corresponding $j^{th}$ element of the ideal vector in each class. The calculation is described mathematically in equation (\ref{eqSim}).

\begin{equation}\label{eqSim}
Sim \langle \vec{x_i}, \vec{v}_k, j \rangle  = \sqrt[p]{1 - |\vec{x_i}(j)^p - \vec{v}_k(j)^p|}
\end{equation}

$p$ is a hyper parameter which is optimized in section \ref{expCombinations}. A similarity value is calculated for each feature in each class for each subject. Subsequently, a $MN \times D$ similarity matrix $\mathbf{P}$ is constructed as shown in Table \ref{similarityMatrix}. The fuzzy entropy calculation for each feature is described in the next subsection.

\begin{table}[h]\scriptsize
    \centering
	\caption{Similarities matrix}
	\begin{tabular}{c c c c c}
		\toprule
		\textbf{Data} & \textbf{Feature 1} & \textbf{Feature 2} & \textbf{...} & \textbf{Feature D}\\
		\midrule
		\centering
		$\vec{x_1}$	 & $Sim \langle \vec{x_1}, \vec{v_1}, 1 \rangle$ &  $Sim \langle \vec{x_1}, \vec{v_1}, 2 \rangle$  & ...&  $Sim \langle \vec{x_1}, \vec{v_1}, D \rangle$  \\
		$\vec{x_1}$	 & $Sim \langle \vec{x_1}, \vec{v_2}, 1 \rangle$ &  $Sim \langle \vec{x_1}, \vec{v_2}, 2 \rangle$  & ...&  $Sim \langle \vec{x_1}, \vec{v_2}, D \rangle$  \\
		...			&  ...  &   ...  &  ...  &  ...\\
		$\vec{x_1}$	 & $Sim \langle \vec{x_1}, \vec{v_N}, 1 \rangle$   &  $Sim \langle \vec{x_1}, \vec{v_N}, 2 \rangle$  & ...&  $Sim \langle \vec{x_1}, \vec{v_N}, D \rangle$  \\
		$\vec{x_2}$	 & $Sim \langle \vec{x_2}, \vec{v_1}, 1 \rangle$    &  $Sim \langle \vec{x_2}, \vec{v_1}, 2 \rangle$  & ...&  $Sim \langle \vec{x_2}, \vec{v_1}, D \rangle$  \\	
		...			&  ...  &   ...  &  ...  &  ...\\
		$\vec{x_M}$	 & $Sim \langle \vec{x_M}, \vec{v_N}, 1 \rangle$   &  $Sim \langle \vec{x_M}, \vec{v_N}, 2 \rangle$  & ...&  $Sim \langle \vec{x_M}, \vec{v_N}, D \rangle$  \\			
		\bottomrule
	\end{tabular}
	\label{similarityMatrix}
\end{table}

\subsection{\textbf{Fuzzy entropy based feature selection}}\label{fuzzyEntropy}
In order to reduce the dimensionality and discard the non-important features, the fuzzy entropy based feature selection process \cite{luukka2011feature} is used to rank the features. Fuzzy entropy is the basic definition of the fuzzy information process and widely used to measure the degree of vagueness among various areas \cite{kosko1986fuzzy}.

Based on the previously constructed similarity matrix, we calculate the fuzzy entropy value for each feature (each column of matrix $\mathbf{P}$) using the fuzzy entropy functions described below. $\mathbf{P}(r,j)$ is used to represent the value of the $r^{th}$ row and $j^{th}$ column in the similarity matrix. These similarity values are utilized as the membership function of fuzzy set in the fuzzy entropy calculation. Three different fuzzy entropy methods are implemented as expressed in equation (\ref{eq1}), (\ref{eq2}) and (\ref{eq3}).

\paragraph{Non Probabilistic Entropy (Luca's method)}
De Luca and Termini \cite{de1972definition} axiomatized non-probabilistic fuzzy entropy functions and defined a fuzzy entropy measurement based on Shannon's entropy as below.
\begin{equation}\label{eq1}
\begin{split}
H_1(j) =& - \sum_{r = 1}^{MN} [(\mathbf{P}(r,j) \log \mathbf{P}(r,j)) \\
&+ (1 - \mathbf{P}(r,j))\log (1 - \mathbf{P}(r,j))]
\end{split}
\end{equation}

\paragraph{Weighted Measurement of Fuzzy Entropy (Parkash's method)}
Parkash \cite{parkash2008new} proposed a new measurement of fuzzy entropy as in equation (\ref{eq2}).

\begin{equation}\label{eq2}
    H_2(j) = \sum_{r = 1}^{MN} sin\frac{\pi \mathbf{P}(r,j)}{2} + sin\frac{\pi(1- \mathbf{P}(r,j))}{2} - 1
\end{equation}

\paragraph{Geometry of Fuzzy Set and Entropy (Kosko's method)}
Kosko \cite{kosko1986fuzzy} utilized the concepts of overlap and underlap to define the fuzzy entropy based on the geometry of hypercube:

\begin{equation}\label{eq3}
H_3(j)= \frac {\sum_{r = 1}^{MN}(\mathbf{P}(r,j) \land (1-\mathbf{P}(r,j)) )}
{\sum_{r = 1}^{MN}(\mathbf{P}(r,j) \lor (1-\mathbf{P}(r,j)))}
\end{equation}

Subsequently the fuzzy entropy values are used for feature ranking and selection. We then perform classification based on the selected features in the next subsection.

\subsection{\textbf{Classification based on the selected features}}\label{classification}
The classification method is based on the maximal fuzzy similarity measures proposed in \cite{luukka2001classifier}. Corresponding to the three methods for idea vector calculation, three similarity measurements are implemented here.

\setcounter{paragraph}{0}
\paragraph{Similarity measure based on arithmetic mean}
\begin{equation}\label{eqAri}
S^A \langle \vec{x_i}, \vec{v}_k \rangle  = \frac{1}{D'} \sum_{j=1}^{D'} \sqrt[p]{1 - |\vec{x_i}(j)^p - \vec{v}_k(j)^p|}
\end{equation}

\paragraph{Similarity measure based on geometric mean}
\begin{equation}\label{eqGeo}
S^G \langle \vec{x_i}, \vec{v}_k \rangle = \sqrt[D']{\prod_{j=1}^{D'} \sqrt[p]{1 - |\vec{x_i}(j)^p - \vec{v}_k(j)^p|}  }
\end{equation}

\paragraph{Similarity measure based on harmonic mean}
\begin{equation}\label{eqHar}
S^H \langle \vec{x_i}, \vec{v}_k \rangle = \frac{D'}{ \sum_{j =1}^{D'} \frac{1}{ \sqrt[p]{1 - |\vec{x_i}(j)^p - \vec{v}_k(j)^p|}}}
\end{equation}

In equation (\ref{eqAri}), (\ref{eqGeo}) and (\ref{eqHar}), $\vec{x_i}$ represents the feature vector of the $i^{th}$ subject in testing set after feature selection. $\vec{v}_k(j)$ stands for the recalculated ideal vector with the reduced dimension in the training set. $D'$ is the number of the selected features. The parameter $p$ is the same as in equation (\ref{eqSim}). Each testing subject is then classified into the class that produces the highest similarity value. It is noteworthy to mention that, based on the reduced feature subset, other classifiers can also be applied and compared, e.g. random forest, support vector machine etc. The comparison of different classifiers is not the main focus of this paper.

\section{Experiments}
\subsection{\textbf{Materials}}
The proposed method was tested on three publicly available biomedical datasets with binary classifications. Those widely tested datasets were all extracted from real clinical problems with different sample to feature ratios. The key properties of those datasets are shown in Table \ref{descriptionData}.

\begin{table}[h]
	\caption{Description of the biomedical datasets}
	\begin{tabular}{l l l l}
		\toprule
		\textbf{Dataset} & \textbf{Nb. Features} & \textbf{Nb. Samples}  & \textbf{Samples/Features}\\
		\midrule
		\centering
		WBC	        		          						  &   9  &  699  &   77.719   \\
		WDBC              									&   31  &  569  &  18.4    \\
		Parkinsons 								  			 &  22  &  197  &   9.0    \\
		\bottomrule
	\end{tabular}
	\label{descriptionData}
\end{table}

\subsubsection{Wisconsin Breast Cancer (WBC)}
Wisconsin breast cancer dataset was generated by Dr. Wolberg from his clinical cases. For data preprocessing, the sample code ID and the rows with nan values were removed. Then the number of samples became 683 after the preprocessing. Nine visually assessed features were considered to predict benign or malignant \cite{wolberg1990multisurface}.

\subsubsection{Wisconsin Diagnostic Breast Cancer (WDBC)}
The features in Wisconsin Diagnostic Breast Cancer dataset were computed from a digitized image of a fine needle aspirate of a breast mass. The features described characteristics of the cell nuclei presented in the image \cite{Lichman2013}. 

\subsubsection{Parkinsons}
The dataset, created by Max Little at the University of Oxford, is composed of a range of biomedical voice measurements from healthy people and the people with Parkinson's disease (PD). The main aim of this data is to discriminate healthy people from those with PD \cite{little2009suitability}.

\begin{figure*}[htbp]
	\centering
	\begin{minipage}[t]{0.3\linewidth}
		\centering
		\includegraphics[width=1.2\textwidth]{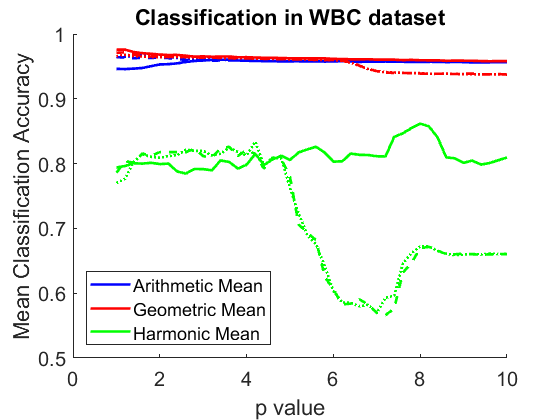}
		\parbox{1cm}{\small \hspace{4cm}(a)WBC}
	\end{minipage}
	\hspace{3ex}   %%两个minipage之间相隔3个字符的距离
	\begin{minipage}[t]{0.3\linewidth}
		\centering
		\includegraphics[width=1.2\textwidth]{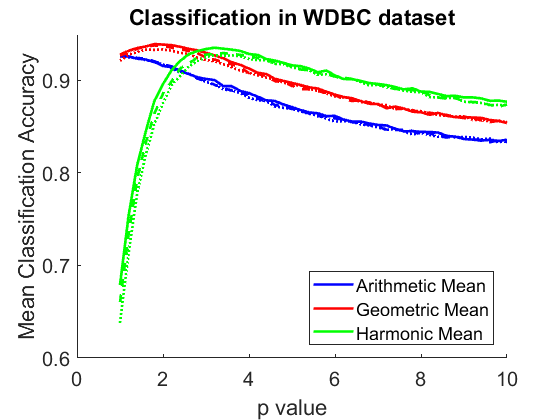}
		\parbox{1cm}{\small \hspace{3.5cm}(b)WDBC}
	\end{minipage}
	\hspace{3ex}   %%两个minipage之间相隔3个字符的距离
	\begin{minipage}[t]{0.3\linewidth}
		\centering
		\includegraphics[width=1.2\textwidth]{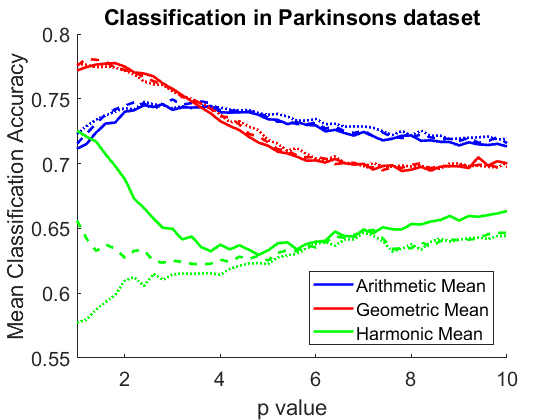}
		\parbox{1cm}{\small \hspace{3.5cm}(c)Parkinsons}
	\end{minipage}
	\caption{Mean classification accuracies with different $p$ values on the different datasets}
	\label{meanAccComparison}
\end{figure*}

\subsection{\textbf{Evaluation of different combinations of ideal vector calculation and classification methods}}\label{expCombinations}
The combination of using three ideal vector calculations and three similarity functions for classification were colour coded and listed in Table \ref{combinations}. In this experiment, the full sets of features were used for both training and testing without performing feature selection. Without affecting by the feature selection results, this allows a fair comparison of different ideal vector calculations combined with different classification methods, as well as $p$ value (equation (\ref{eqSim})) optimization. The methods were tested and compared on three biomedical datasets by evaluating the classification accuracy. The classification accuracy was defined as the number of correctly classified subjects divided by the total number of subjects.

\begin{table}[h]
	\centering
	\caption{Different classifier combinations used}
	\label{combinations}
	\begin{tabular}{|c|c|c|c|}
		\hline
		\textbf{Ideal vector} & \textbf{Classification methods} & \textbf{Name}  & \textbf{Line}  \\ \hline
		\multirow{3}{*}{Arithmetic mean}   &  Arithmetic mean    & A-A   &  \includegraphics[width=0.02\textwidth]{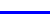}  \\ \cline{2-4}
		&  Geometric mean   &  A-G  &  \includegraphics[width=0.02\textwidth]{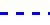}   \\ \cline{2-4}
		&  Harmonic mean   &   A-H   &  \includegraphics[width=0.02\textwidth]{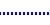} \\ \hline
		\multirow{3}{*}{Geometric mean} 	  &   Arithmetic mean  &   G-A &  \includegraphics[width=0.02\textwidth]{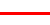}  \\  \cline{2-4}
		&   Geometric mean &    G-G  &  \includegraphics[width=0.02\textwidth]{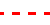}  \\  \cline{2-4}
		&   Harmonic mean  &    G-H   &  \includegraphics[width=0.02\textwidth]{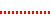} \\  \hline
		\multirow{3}{*}{Harmonic mean}  &   Arithmetic mean  &   H-A  &  \includegraphics[width=0.02\textwidth]{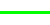}  \\  \cline{2-4}
		&   Geometric mean  &   H-G  &  \includegraphics[width=0.02\textwidth]{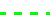}  \\  \cline{2-4}
		&   Harmonic mean    &  H-H  &  \includegraphics[width=0.02\textwidth]{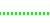} \\  \hline
	\end{tabular}
\end{table}

Same as the evaluation in Luukka's work \cite{luukka2011feature}, the datasets were divided into two halves. One half was used for training and the other half for testing. Additional to the experiment in \cite{luukka2011feature}, we also repeated the experiments for 1000 times for each $p$ value (equation (\ref{eqSim})) with random two-half group splitting. Note that all the remaining experiments in this paper for classification accuracy calculation were tested based on the same evaluation mechanism, if not explicitly described. The mean classification accuracy of those combinations on the three biomedical datasets were evaluated and presented in Fig. \ref{meanAccComparison}.

Fig. \ref{meanAccComparison}-a shows the results of the mean classification accuracies for the WBC dataset. In this case, the idea vector calculation using the arithmetic and geometric mean have achieved similar results, which are much higher and more stable than using the harmonic mean. The curves of using the harmonic mean methods vary dramatically when $p$ value is greater than 5.

Fig. \ref{meanAccComparison}-b shows the mean classification accuracies for WDBC dataset at different $p$ values. It is seen that different classification methods with the same ideal vector method achieved similar performances. The ideal vector calculation using arithmetic mean and geometric mean methods decreased slowly when $p$ value increased. However, in the case of harmonic mean method, the accuracies increased sharply and peaked at $p=3$. Then the mean classification accuracies decreased slowly along with the other two ideal vector calculation methods.

Fig. \ref{meanAccComparison}-c presents the results for the Parkinsons dataset. Arithmetic mean method for calculating the idea vector produced quite stable mean classification accuracy around 0.73. The accuracies of the methods using geometric mean for ideal vector calculation were maximized at the value around 0.78 and then dropped down quickly when $p$ was greater than 2. The harmonic mean methods for ideal vector calculation produced the worst and unstable results.

Overall, the geometric mean method for calculating the idea vector have achieved the maximal classification accuracies when $p$ is around 2 for all of the three datasets. There are not much differences by using the three different similarity functions for classification. Therefore in the following experiments, geometric mean methods were utilized for both ideal vector calculation and maximal similarity classification. The $p$ value in equation (\ref{eqSim}) and (\ref{eqGeo}) was set to be 2.

\begin{figure*}[h]
	\centering
	\begin{minipage}[t]{0.3\linewidth}
		\centering
		\includegraphics[width=1.2\textwidth]{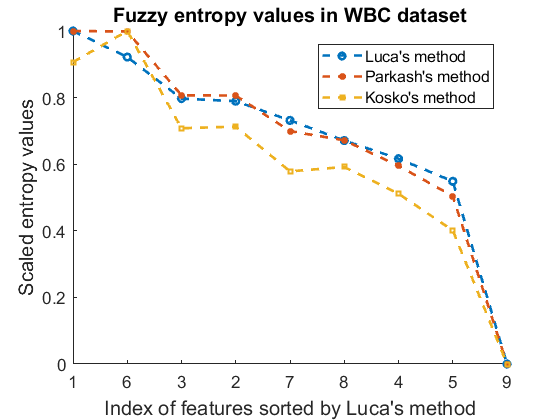}
		\parbox{1cm}{\small \hspace{4.5cm}(a)WBC}
	\end{minipage}
	\hspace{3ex}
	\begin{minipage}[t]{0.3\linewidth}
		\centering
		\includegraphics[width=1.2\textwidth]{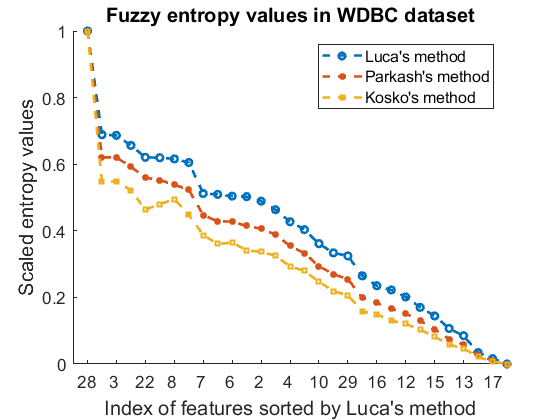}
		\parbox{1cm}{\small \hspace{3.5cm}(b)WDBC}
	\end{minipage}
	\hspace{3ex}
	\begin{minipage}[t]{0.3\linewidth}
		\centering
		\includegraphics[width=1.2\textwidth]{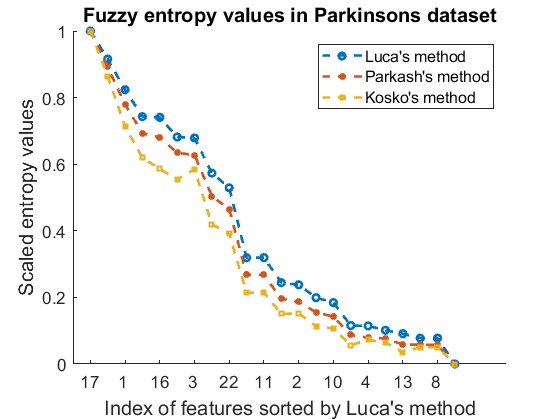}
		\parbox{1cm}{\small \hspace{3.5cm}(c)Parkinsons}
	\end{minipage}
	\caption{Scaled entropy values of the sorted features for different datasets}
	\label{entropyFeatureComparison}
\end{figure*}

\begin{figure*}[h]%[htbp]
	\centering
	\begin{minipage}[t]{0.3\linewidth}
		\centering
		\includegraphics[width=1.2\textwidth]{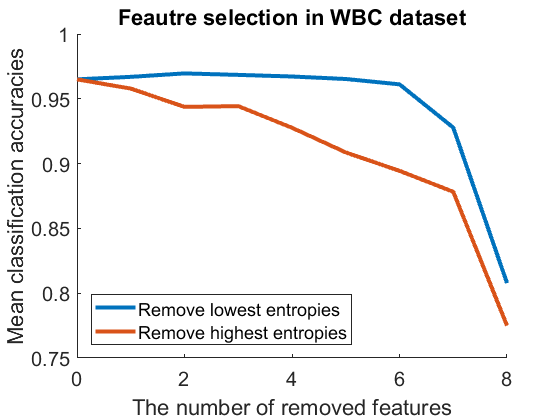}
		\parbox{1cm}{\small \hspace{3.5cm}(a)WBC}
	\end{minipage}
	\hspace{3ex}   %%两个minipage之间相隔3个字符的距离
	\begin{minipage}[t]{0.3\linewidth}
		\centering
		\includegraphics[width=1.2\textwidth]{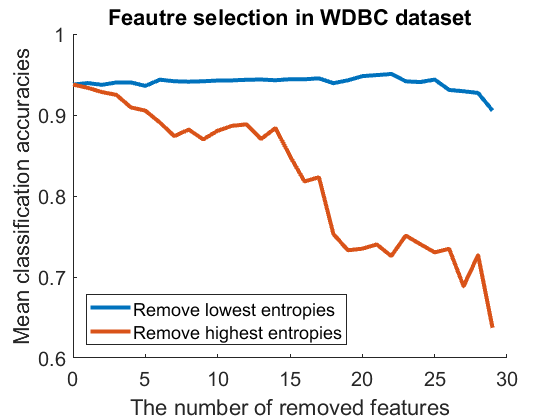}
		\parbox{1cm}{\small \hspace{3.5cm}(b)WDBC}
	\end{minipage}
	\hspace{3ex}   %%两个minipage之间相隔3个字符的距离
	\begin{minipage}[t]{0.3\linewidth}
		\centering
		\includegraphics[width=1.2\textwidth]{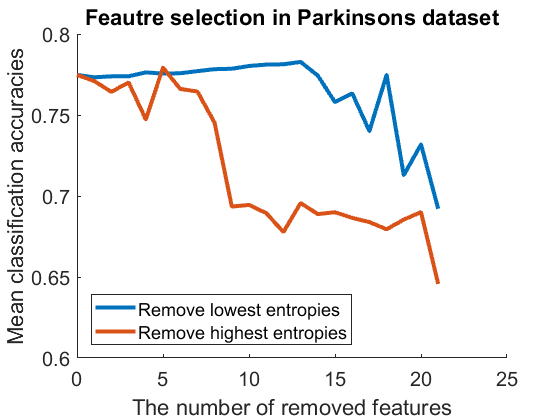}
		\parbox{1cm}{\small \hspace{3.5cm}(c)Parkinsons }
	\end{minipage}
	\caption{Comparison of the different feature removing orders based on fuzzy entropy values}
	\label{RemovingOrder}
\end{figure*}

\subsection{\textbf{Evaluation of different fuzzy entropy methods}}
The aim of this experiment was to compare the feature ranking sequence produced by three different fuzzy entropy methods in section \ref{fuzzyEntropy}. The fuzzy entropy values were used to rank the features from high to low. In order to compare different methods, the entropy values were then normalized to the range of [0 1] by min-max normalization. For ease of comparison, we randomly chose one method (Luca's method in this section) as the reference ranking sequence in the horizontal axis of Fig. \ref{entropyFeatureComparison}. All the indices of features were sorted according to the reference ranking.

It is observed from Fig. \ref{entropyFeatureComparison} that different fuzzy entropy functions produced similar ranking sequences for the three datasets. Luca's method and Parkash's method resulted in almost identical ranking sequence for all the datasets. The result from Kosko's method showed disagreement at multiple points with the other two, especially in Parkinsons dataset. 

According to our experiment results, the ranking differences of the three methods did not make a significant impact on the final classification performance. We chose Luca's method for the fuzzy entropy function in our final framework, as it produced the highest consistency with the other two methods. 

\subsection{\textbf{Evaluation on different removing order of fuzzy entropy methods}}
In order to explore the optimal feature selection process, we also compared different feature removal order according to the entropy values. Two different feature selection approaches were compared. One method removed the feature with the highest entropy value each time. The other method removed the feature with the lowest entropy value each time. The mean classification accuracies using the two different feature removal orders for the three datasets are shown in Fig. \ref{RemovingOrder}.

It is observed from Fig. \ref{RemovingOrder} that the method that removed the feature with the lowest entropy value each time maintained a high performance even when half of the features were removed for all three datasets. In contrast, as soon as one feature with the highest entropy value was removed, the performance dropped significantly for all three datasets. Therefore, we concluded that the feature selection approach that eliminated the feature with the lowest entropy value each time should be used. However, this conclusion is contradictory with Luukka's suggestion in \cite{luukka2011feature}, which removes the feature with the highest fuzzy entropy value each time.

\begin{figure*}[h]%[htbp]
	\centering
	\begin{minipage}[t]{0.3\linewidth}
		\centering
		\includegraphics[width=1.2\textwidth]{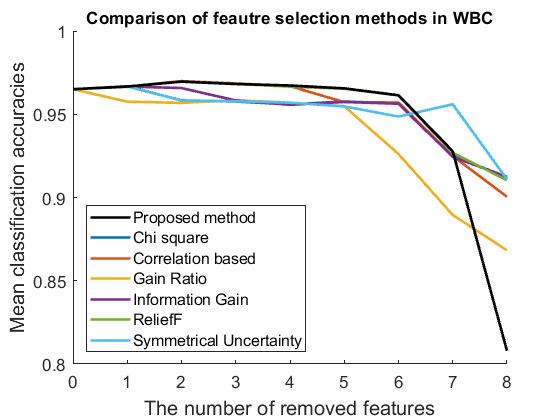}
		\parbox{1cm}{\small \hspace{4.5cm}(a)WBC}
	\end{minipage}
	\hspace{3ex}   %%两个minipage之间相隔3个字符的距离
	\begin{minipage}[t]{0.3\linewidth}
		\centering
		\includegraphics[width=1.2\textwidth]{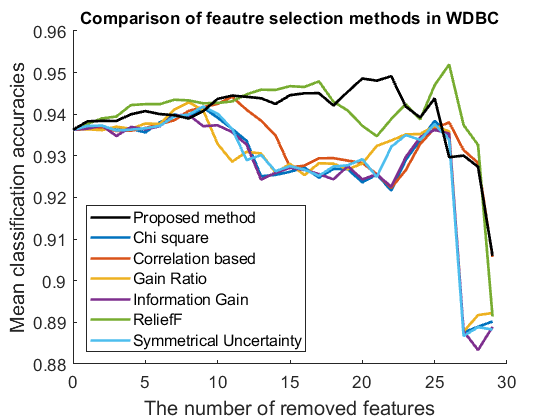}
		\parbox{1cm}{\small \hspace{3.5cm}(b)WDBC}
	\end{minipage}
	\hspace{3ex}   %%两个minipage之间相隔3个字符的距离
	\begin{minipage}[t]{0.3\linewidth}
		\centering
		\includegraphics[width=1.2\textwidth]{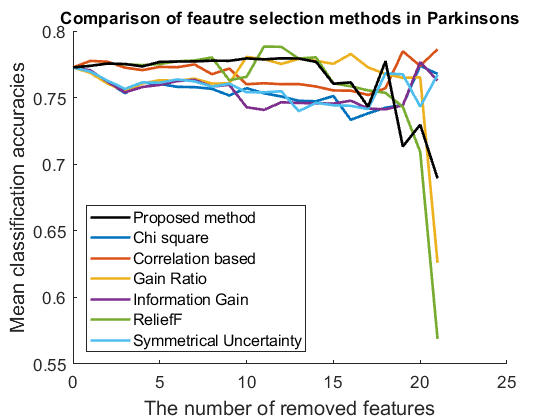}
		\parbox{1cm}{\small \hspace{3.5cm}(c)Parkinsons}
	\end{minipage}
	\caption{Comparison of different feature selection methods on different datasets}
	\label{FSComparison}
\end{figure*}

\subsection{\textbf{Comparison with other feature selection methods}}
Based on the previous experiments, we have found the optimized combination of different methods within the proposed feature selection and classification framework. The optimal choice and settings are: geometric mean method for ideal vector calculation and classification function with $p=2$ and Luca's method for fuzzy entropy calculation. In this section, we compared the proposed method with other six classical filter based feature selection methods in the literature including Chi square based \cite{jin2006machine}, Correlation based \cite{hall1999correlation} , Gain ratio based \cite{karegowda2010comparative}, Information gain based \cite{lee2006information}, ReliefF based \cite{liu2007computational} and Symmetrical uncertainty based \cite{yu2004efficient}. All the filter based feature selection methods rank the features from higher to lower values. The same maximal similarity classifier with geometric method was used for the classification task after the feature selection process for all the compared methods. The mean classification accuracies of the different feature selection methods on three datasets are presented in Fig. \ref{FSComparison}.

Fig.\ref{FSComparison}-(a) has shown that in WBC dataset, the proposed method maintains the highest classification accuracies among all the methods with the number of removed features increasing from 0 to 6. In WDBC dataset (Fig.\ref{FSComparison}-(b)), the top two performers are the proposed method and ReliefF method. The classification accuracies keep increasing even when about 20 features have been removed by using the proposed method. For Parkinsons dataset (Fig.\ref{FSComparison}-(c)), the proposed method maintained a stable performance with arguably the highest classification accuracies until 14 features being removed.

Another important observation is that the classification accuracies of the proposed method generally follow the trend of gradually increasing, achieving peak performance and decreasing when features were gradually removed for all three datasets. This is a good indication of the features were ranked, from the least to the most importance, reasonably well. However, the performances of other methods changed dramatically when features were gradually removed (Fig. \ref{FSComparison}-(b) and Fig. \ref{FSComparison}-(c)).

For performance comparison of feature selection methods, it has not been standardized in the literature. One option is to report the highest classification accuracy despite the number of features selected. Alternatively, the classification accuracies are compared based on the same number of selected features. Arguably, if the classification result is more important, the first option should be applied. In this paper, we aims to compare different feature selection methods, where the compactness, representative and relevance of the selected features are more important in this case. Therefore, we adopted the second option and proposed the following comparison criteria. 

\begin{table}[h]
	\centering
	\caption{Mean classification accuracy of different feature selection methods}
	\label{comFS}
	\begin{tabular}{|c|c|c|c|c|c|c|}
		\hline
		\multirow{2}{*}{Methods} 		& \multicolumn{2}{c|}{WBC} & \multicolumn{2}{c|}{WDBC} & \multicolumn{2}{c|}{Parkinsons} \\ \cline{2-7}
		& Acc.(\%)   &   Nb.        &   Acc. (\%) &  Nb.   & Acc. (\%)  &  Nb.  \\ \hline
Proposed		&  \textbf{96.97} &  \textbf{7}      &   94.86     &    8       &       78.23    &    9       \\ \hline
Chi square		&  95.86 &     7                 &    93.67       &   5        &      77.70     &    2       \\ \hline
Correlation		&   96.95 &    7                 &   93.86        &  4         &    \textbf{78.72}       &    \textbf{3}       \\ \hline
Gain Ratio		&   95.83 &   6                 &   93.73        &    5       &     78.09      &    6       \\ \hline
Info. Gain		&    96.53 &    7                &    93.71       &    5       &     77.43      &      2     \\ \hline
ReliefF		     &     96.96  &   7              &   \textbf{95.21}  &  \textbf{4}     &  78.26       &   8    \\ \hline
Sym. Unc.	&   95.84 &    7                  &      93.68     &     5      &   77.19        &     3     \\ \hline
	\end{tabular}
\end{table}

We chose the proposed method as the reference method to compare with each of the other competitors. The selected number of features (denoted as $S$) that produced the highest mean classification accuracy of our method was used as the reference. For other methods, the highest mean classification accuracies were reported with the selected number of features less or equal to $S$. For comparison, higher classification accuracy indicates better feature selection performance. Additionally, McNemar's test \cite{bennett1970283} was applied to test the statistical significance of the binary classification results for each of the two compared methods.

\begin{table}[h]
	\centering
	\caption{$P$ values of McNemar's test for the pairwise tests between the proposed method and each of the competitors}
	\label{Pvalue}
	\begin{tabular}{|c|c|c|c|}
		\hline
		Methods  		&     \textbf{WBC}              &      \textbf{WDBC}          &      \textbf{Parkinsons}          \\ \hline
		Chi square		&    $<$0.001             &     $<$0.001             &   $<$0.001       \\ \hline
		Correlation	  	&     1.000                  &     $<$0.001             &    $<$0.001       \\ \hline
		Gain Ratio		&     $<$0.001                 &     $<$0.001              &    $<$0.001       \\ \hline
		Info. Gain		 &    $<$0.001            &     $<$0.001              &    $<$0.001       \\ \hline
		ReliefF		      &     1.000                  &     $<$0.001             &     $<$0.001      \\ \hline
		Sym. Unc.	   &     $<$0.001            &     $<$0.001             &     $<$0.001      \\ \hline
	\end{tabular}
\end{table}

The mean classification accuracies (Acc.) and the selected number of features (Nb.) for the three datasets are listed in Table \ref{comFS}. The $P$ values of McNemar's test for the pairwise tests between the proposed method and each of the competitors are presented in Table \ref{Pvalue}. For the results of the WBC dataset in Table IV, it is observed that the proposed method produced the best mean classification accuracy compared with other methods with $S=7$. According to Table V, it is statistically better than the Chi square, Gain Ratio, Info. Gain and Sym. Unc. methods but no statistical differences to the Correlation and ReliefF methods. 

For WDBC dataset, the proposed method produced the second best classification accuracy with $S=8$. Each of other methods achieved individually higher performance of using about 4 or 5 features rather than 8 features. However, from Fig. 5-(b), it is seen that the proposed method was still the second best, if 5 features were used (the value corresponding to 26 in the horizontal axes of Fig. 5-(b)). According to Table V, the proposed method is statistically worse than the ReliefF method, but statistically better than all other methods. 

For Parkinsons dataset, the proposed method ($S=9$) ranked the third, which was statistically worse than the Correlation (3 features) and RefliefF methods (8 features). The other methods were statistically worse than the proposed method.

\section{Discussion \& Conclusion}
In this paper, based on Luukka's \cite{luukka2011feature} fuzzy entropy feature selection framework, we have implemented and compared different methods within each of the key components of the framework. They include the combinations of using three ideal vector calculations, three maximal similarity classifiers and three fuzzy entropy functions. All the evaluations were performed on three widely used publicly available biomedical datasets. All these three datasets were generated from challenging clinical applications with different feature to subject ratios. All experiments were thoroughly tested by evenly and randomly splitting the dataset into a training and a testing group, and repeated for 1000 times. From the experiments, we found that the use of geometric method for ideal vector calculation ($p=2$), geometric method for similarity classifier ($p=2$) and Luca's method for fuzzy entropy calculation achieved the most stable performance and highest classification accuracy. Additionally, we concluded that features with the lowest entropy value should be removed each time to achieve the best performance. 

We further compared the proposed method with other six classical filter-based feature selection methods. The mean classification accuracies were compared by fixing the number of selected features. McNemar's test was also applied to evaluate the statistical differences between the pairwise comparisons. The proposed method produced the highest classification accuracy for WBC dataset, ranked the $2^{nd}$ best and $3^{rd}$ best for WDBC and Parkinsons datasets respectively. Correlation method, ReliefF method and the proposed method are the top performers among the compared methods. More importantly, from the results, it is shown that the proposed method achieved the most stable performance for all three datasets when the features being gradually removed. This indicates a better feature ranking performance.

For future work, we will test our method on different datasets for various applications with more features and more subjects. The robustness of the proposed method that handles outliers and incomplete data will be investigated.

% \section*{Acknowledgment}

%----------------------------------------------------------------------------------------
%	BIBLIOGRAPHY
%----------------------------------------------------------------------------------------
\bibliographystyle{IEEEtran}
\bibliography{sample.bib}

\begin{thebibliography}{10}
\providecommand{\url}[1]{#1}
\csname url@samestyle\endcsname
\providecommand{\newblock}{\relax}
\providecommand{\bibinfo}[2]{#2}
\providecommand{\BIBentrySTDinterwordspacing}{\spaceskip=0pt\relax}
\providecommand{\BIBentryALTinterwordstretchfactor}{4}
\providecommand{\BIBentryALTinterwordspacing}{\spaceskip=\fontdimen2\font plus
\BIBentryALTinterwordstretchfactor\fontdimen3\font minus
  \fontdimen4\font\relax}
\providecommand{\BIBforeignlanguage}[2]{{%
\expandafter\ifx\csname l@#1\endcsname\relax
\typeout{** WARNING: IEEEtran.bst: No hyphenation pattern has been}%
\typeout{** loaded for the language `#1'. Using the pattern for}%
\typeout{** the default language instead.}%
\else
\language=\csname l@#1\endcsname
\fi
#2}}
\providecommand{\BIBdecl}{\relax}
\BIBdecl

\bibitem{kumar2014feature}
V.~Kumar and S.~Minz, ``Feature selection,'' \emph{SmartCR}, vol.~4, no.~3, pp.
  211--229, 2014.

\bibitem{domingos1999role}
P.~Domingos, ``The role of occam's razor in knowledge discovery,'' \emph{Data
  mining and knowledge discovery}, vol.~3, no.~4, pp. 409--425, 1999.

\bibitem{kononenko2001machine}
I.~Kononenko, ``Machine learning for medical diagnosis: history, state of the
  art and perspective,'' \emph{Artificial Intelligence in medicine}, vol.~23,
  no.~1, pp. 89--109, 2001.

\bibitem{bellman2013dynamic}
R.~Bellman, \emph{Dynamic programming}.\hskip 1em plus 0.5em minus 0.4em\relax
  Courier Corporation, 2013.

\bibitem{li2016feature}
J.~Li, K.~Cheng, S.~Wang, F.~Morstatter, R.~P. Trevino, J.~Tang, and H.~Liu,
  ``Feature selection: A data perspective,'' \emph{arXiv preprint
  arXiv:1601.07996}, 2016.

\bibitem{bandemer2012fuzzy}
H.~Bandemer and W.~N{\"a}ther, \emph{Fuzzy data analysis}.\hskip 1em plus 0.5em
  minus 0.4em\relax Springer Science \& Business Media, 2012, vol.~20.

\bibitem{rezaee1999fuzzy}
M.~R. Rezaee, B.~Goedhart, B.~P. Lelieveldt, and J.~H. Reiber, ``Fuzzy feature
  selection,'' \emph{Pattern Recognition}, vol.~32, no.~12, pp. 2011--2019,
  1999.

\bibitem{li2002fuzzy}
R.-P. Li, M.~Mukaidono, and I.~B. Turksen, ``A fuzzy neural network for pattern
  classification and feature selection,'' \emph{Fuzzy Sets and Systems}, vol.
  130, no.~1, pp. 101--108, 2002.

\bibitem{tsang2008attributes}
E.~C. Tsang, D.~Chen, D.~S. Yeung, X.-Z. Wang, and J.~W. Lee, ``Attributes
  reduction using fuzzy rough sets,'' \emph{IEEE Transactions on Fuzzy
  systems}, vol.~16, no.~5, pp. 1130--1141, 2008.

\bibitem{luukka2011feature}
P.~Luukka, ``Feature selection using fuzzy entropy measures with similarity
  classifier,'' \emph{Expert Systems with Applications}, vol.~38, no.~4, pp.
  4600--4607, 2011.

\bibitem{Lichman2013}
\BIBentryALTinterwordspacing
M.~Lichman, ``{UCI} machine learning repository,'' 2013. [Online]. Available:
  \url{http://archive.ics.uci.edu/ml}
\BIBentrySTDinterwordspacing

\bibitem{jain2011min}
Y.~K. Jain and S.~K. Bhandare, ``Min max normalization based data perturbation
  method for privacy protection,'' \emph{International Journal of Computer \&
  Communication Technology}, vol.~2, no.~8, pp. 45--50, 2011.

\bibitem{saastamoinen2003testing}
K.~Saastamoinen and P.~Luukka, ``Testing continuous t-norm called lukasiewicz
  algebra with different means in classification,'' in \emph{Fuzzy Systems,
  2003. FUZZ'03. The 12th IEEE International Conference on}, vol.~2.\hskip 1em
  plus 0.5em minus 0.4em\relax IEEE, 2003, pp. 808--813.

\bibitem{kosko1986fuzzy}
B.~Kosko, ``Fuzzy entropy and conditioning,'' \emph{Information sciences},
  vol.~40, no.~2, pp. 165--174, 1986.

\bibitem{de1972definition}
A.~De~Luca and S.~Termini, ``A definition of a nonprobabilistic entropy in the
  setting of fuzzy sets theory,'' \emph{Information and control}, vol.~20,
  no.~4, pp. 301--312, 1972.

\bibitem{parkash2008new}
O.~Parkash, P.~Sharma, and R.~Mahajan, ``New measures of weighted fuzzy entropy
  and their applications for the study of maximum weighted fuzzy entropy
  principle,'' \emph{Information Sciences}, vol. 178, no.~11, pp. 2389--2395,
  2008.

\bibitem{luukka2001classifier}
P.~Luukka, K.~Saastamoinen, and V.~Kononen, ``A classifier based on the maximal
  fuzzy similarity in the generalized lukasiewicz-structure,'' in \emph{Fuzzy
  Systems, 2001. The 10th IEEE International Conference on}, vol.~1.\hskip 1em
  plus 0.5em minus 0.4em\relax IEEE, 2001, pp. 195--198.

\bibitem{wolberg1990multisurface}
W.~H. Wolberg and O.~L. Mangasarian, ``Multisurface method of pattern
  separation for medical diagnosis applied to breast cytology.''
  \emph{Proceedings of the national academy of sciences}, vol.~87, no.~23, pp.
  9193--9196, 1990.

\bibitem{little2009suitability}
M.~A. Little, P.~E. McSharry, E.~J. Hunter, J.~Spielman, L.~O. Ramig
  \emph{et~al.}, ``Suitability of dysphonia measurements for telemonitoring of
  parkinson's disease,'' \emph{IEEE transactions on biomedical engineering},
  vol.~56, no.~4, pp. 1015--1022, 2009.

\bibitem{jin2006machine}
X.~Jin, A.~Xu, R.~Bie, and P.~Guo, ``Machine learning techniques and chi-square
  feature selection for cancer classification using sage gene expression
  profiles,'' in \emph{International Workshop on Data Mining for Biomedical
  Applications}.\hskip 1em plus 0.5em minus 0.4em\relax Springer, 2006, pp.
  106--115.

\bibitem{hall1999correlation}
M.~A. Hall, ``Correlation-based feature selection for machine learning,'' 1999.

\bibitem{karegowda2010comparative}
A.~G. Karegowda, A.~Manjunath, and M.~Jayaram, ``Comparative study of attribute
  selection using gain ratio and correlation based feature selection,''
  \emph{International Journal of Information Technology and Knowledge
  Management}, vol.~2, no.~2, pp. 271--277, 2010.

\bibitem{lee2006information}
C.~Lee and G.~G. Lee, ``Information gain and divergence-based feature selection
  for machine learning-based text categorization,'' \emph{Information
  processing \& management}, vol.~42, no.~1, pp. 155--165, 2006.

\bibitem{liu2007computational}
H.~Liu and H.~Motoda, \emph{Computational methods of feature selection}.\hskip
  1em plus 0.5em minus 0.4em\relax CRC Press, 2007.

\bibitem{yu2004efficient}
L.~Yu and H.~Liu, ``Efficient feature selection via analysis of relevance and
  redundancy,'' \emph{Journal of machine learning research}, vol.~5, no. Oct,
  pp. 1205--1224, 2004.

\bibitem{bennett1970283}
B.~Bennett and R.~Underwood, ``283. note: On mcnemar's test for the 2 * 2 table
  and its power function,'' \emph{Biometrics}, pp. 339--343, 1970.

\end{thebibliography}

\end{document}